\documentclass[sigconf]{acmart}
\AtBeginDocument{%
  }

\copyrightyear{2026}
\acmYear{2026}
\setcopyright{cc}
\setcctype{by}

\acmConference[MM '26]
  {Proceedings of the 34th ACM International Conference on Multimedia}
  {November 10--14, 2026}
  {Rio de Janeiro, Brazil.}

\acmBooktitle{Proceedings of the 34th ACM International
  Conference on Multimedia (MM '26), November 10--14,
  2026, Rio de Janeiro, Brazil}

\acmISBN{979-8-4007-2213-4/2026/11}
\acmDOI{10.1145/3767308.3834979}

\acmSubmissionID{mfp0470}

\usepackage{multirow} 
\usepackage{tabularx, booktabs}
\usepackage{graphicx} 
\usepackage[table]{xcolor} 
\usepackage{utfsym}
\usepackage{pifont}
\newcommand{\cmark}{\ding{51}} 
\newcommand{\xmark}{\ding{55}} 
\begin{document}
\title{C2FXNet: Coarse-to-Fine Scene Expert for Unified Object Detection across Adverse Weather}

\author{Tianle Fang}
\authornote{Tianle Fang and Zhenbing Liu contributed equally to this work.}
\orcid{0009-0007-2461-6685}
\affiliation{%
  \department{School of Computer Science and Information Security}
  \institution{Guilin University of Electronic Technology}
  \city{Guilin}
  \state{Guangxi}
  \country{China}
}

\email{polarisftl123@gmail.com}

\author{Zhenbing Liu}
\authornotemark[1]
\orcid{0000-0001-6551-4174}
\affiliation{%
  \department{School of Artificial Intelligence}
  \institution{Guilin University of Electronic Technology}
  \city{Guilin}
  \state{Guangxi}
  \country{China}
}
\email{zbliu@guet.edu.cn}

\author{Chong Yin}
\orcid{0000-0002-2423-2442}
\affiliation{%
  \department{School of Computer Science and Technology}
  \institution{Hainan University}
  \city{Haikou}
  \state{Hainan}
  \country{China}
}
\email{chongyin@comp.hkbu.edu.hk}

\author{Bolun Li}
\orcid{0009-0008-7218-241X}
\affiliation{%
  \department{School of Computer Science and Information Security}
  \institution{Guilin University of Electronic Technology}
  \city{Guilin}
  \state{Guangxi}
  \country{China}
}
\email{blli@mails.guet.edu.cn}

\author{Haoxiang Lu}
\orcid{0000-0003-2284-5154}
\authornote{Corresponding author.}
\affiliation{%
  \department{School of Computer Science and Information Security}
  \institution{Guilin University of Electronic Technology}
  \city{Guilin}
  \state{Guangxi}
  \country{China}
}
\email{hxlu1005@guet.edu.cn}

\renewcommand{\shortauthors}{Tianle Fang, Zhenbing Liu, Chong Yin, Bolun Li, and Haoxiang Lu}

\begin{abstract}
Object detection in adverse weather remains challenging because severe degradations weaken visual quality and disrupt semantic feature representations across diverse scenes. Existing methods usually rely on condition-specific designs, which limits their ability to generalize within a unified detector. In this paper, we propose a Coarse-to-Fine Scene Expert Network (C2FXNet) that achieves unified detection through hierarchical scene guidance. Specifically, C2FXNet introduces a dual-level guidance mechanism consisting of a Multi-step Reasoning Router (MRR), which performs GRU-based recurrent scene reasoning over compressed multi-scale visual cues and frozen coarse scene prototypes, and a Fine Scene Refinement (FSR) module, which uses image-specific semantic cues to modulate high-level features for local variation handling. Furthermore, a Scene-aware Mixture-of-Experts (SMoE) dynamically combines scene-specific experts under the joint guidance of MRR and FSR. By coupling coarse scene reasoning with fine-grained semantic refinement, C2FXNet enables robust multi-scene detection without scene-specific training. Extensive experiments on RTTS, ExDark, and our newly constructed Adverse Weather Dataset (AWD) demonstrate that C2FXNet consistently outperforms state-of-the-art methods across foggy, dark, and clear conditions, reaching 63.70\%, 71.14\%, and 54.19\% mAP on RTTS, ExDark, and AWD, respectively. The source code will be released at \url{https://github.com/PolarisFTL/C2FXNet}.
\end{abstract}


\maketitle

\section{Introduction}
Object detection is a fundamental computer vision task, with critical applications in autonomous driving \cite{auto-obj, auto-obj1} and outdoor surveillance \cite{Surveillance}. However, real-world deployment often encounters adverse weather conditions (fog, rain, etc.), which 
significantly degrade detection performance due to reduced 
visibility, color distortion, and feature ambiguity. Developing 
robust detection systems that adapt to adverse conditions remains challenging.

\begin{figure}
    \centering
    \includegraphics[width=1\linewidth]{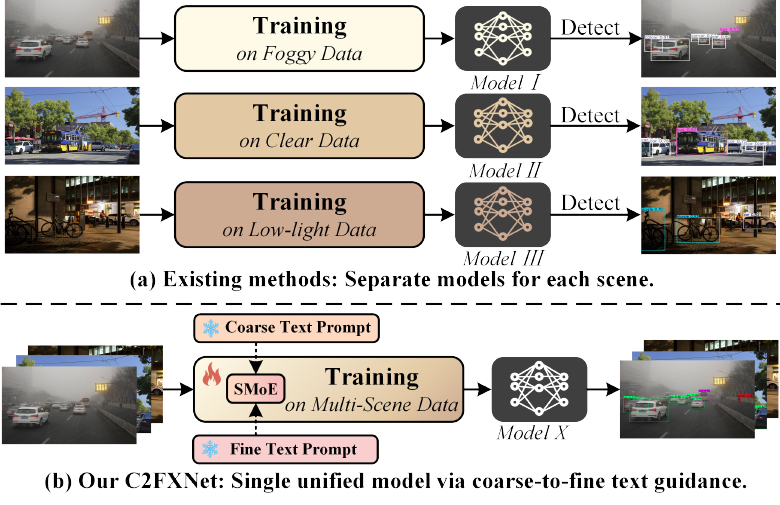}
    \caption{(a) Existing methods: Separate models for each scene.
(b) Our C2FXNet: Single unified model via coarse-to-fine text guidance.}
    \label{fig:placeholder}
\end{figure}

Existing approaches to scene-specific detection mainly follow three
paradigms. Image enhancement methods restore degraded inputs before
detection but may introduce artifacts and disrupt end-to-end
optimization~\cite{dehazing, Degraded, alhindaassi2025adam}.
Multi-task frameworks jointly optimize restoration and detection,
improving robustness at the cost of careful loss
balancing~\cite{dsnet-huang2020, Detection-friendly, CANet}.
Domain adaptation methods transfer detectors from clear to adverse
weather but often remain specialized to particular
conditions~\cite{iayolo-liu2022image, gdip-kalwar2023, masfnet}.
Fundamentally, all three paradigms treat 
different weather conditions as isolated problems rather than 
related tasks sharing common visual semantics. This prevents 
knowledge transfer across conditions and limits generalization to 
unseen weather variations.
Our key insight is that adverse weather conditions, despite their
diverse low-level appearances, share high-level semantic attributes,
such as reduced visibility and contrast, that can be described through
language. Vision-language models such as CLIP~\cite{radford2021clip}
provide transferable semantic priors for modeling these conditions as
related variations rather than isolated domains. However, their global
alignment lacks the hierarchical understanding and region-level
adaptation required for weather-adaptive detection. We therefore employ
two-level textual prompts: coarse scene prompts (foggy, dark, or clear)
guide recurrent scene reasoning and expert routing, while fine prompts
encoding visibility, illumination, and object context refine regional
features. This design enables hierarchical weather adaptation within a
unified detection framework.

Building on this insight, we introduce a coarse-to-fine scene expert 
network (C2FXNet) for unified object detection across adverse weather. 
The framework progressively refines scene understanding from coarse 
global reasoning to fine-grained local adaptation, culminating in 
dynamic expert fusion. Specifically, C2FXNet contains two complementary 
modules aligned with the two-level prompt hierarchy: a Multi-step 
Reasoning Router (MRR) that compresses multi-scale visual cues and 
iteratively updates a latent routing state with GRU units before 
matching it to frozen coarse scene prototypes, and a Fine Scene 
Refinement (FSR) module that uses image-specific semantic prompts to 
refine high-level features for local variations. On top of them, a 
Scene-aware Mixture-of-Experts (SMoE) dynamically combines 
scene-specific experts under joint coarse-to-fine guidance, enabling 
unified detection without manual scene classification. Our 
contributions are summarized as follows:

\begin{itemize}
    \item We propose C2FXNet, a coarse-to-fine scene expert network for unified object detection under adverse weather, which progressively refines scene understanding from coarse scene reasoning to fine-grained local adaptation.
    \item We develop a Multi-step Reasoning Router (MRR) that performs GRU-based recurrent scene reasoning over compressed multi-scale descriptors and frozen coarse scene prototypes, producing sparse and interpretable routing weights with load-balancing regularization.
    \item We introduce Fine Scene Refinement (FSR) to handle intra-condition variations by using image-specific semantic prompts and normalization-based channel-wise modulation with learnable affine parameters.
    \item We devise a Scene-aware Mixture-of-Experts (SMoE) that adaptively fuses scene-specific experts under the joint guidance of MRR and FSR. Extensive experiments on RTTS, ExDark, and AWD validate the effectiveness and robustness of the proposed framework.
\end{itemize}

\section{Related work}

\subsection{Object Detection in Adverse Weather.}
Object detection is a core task in computer vision, which requires both classification and localization. Despite rapid progress, mainstream detectors such as the YOLO series \cite{ scaled-yolov4, yolov7, spikingyolox} suffer performance degradation under adverse weather conditions due to image quality deterioration. Existing solutions can be categorized into three main approaches. Image enhancement methods \cite{dehazing, Degraded, alhindaassi2025adam} restore degraded inputs before detection, but often lack end-to-end design and incur high computational cost. Multi-task learning frameworks \cite{dsnet-huang2020, Detection-friendly, RDFNet} jointly optimize detection and auxiliary tasks (dehazing or low-light enhancement), improving robustness but requiring complex architectures and loss balancing. Domain adaptation techniques \cite{iayolo-liu2022image, gdip-kalwar2023, RDANet} leverage unsupervised transfer to adapt models to adverse weather, but these methods lead to color distortion of the image and detection inconsistency. While these methods mitigate some adverse effects, challenges remain in feature degradation and limited generalization across scenes.

\begin{figure*}
    \centering
    \includegraphics[width=1\linewidth]{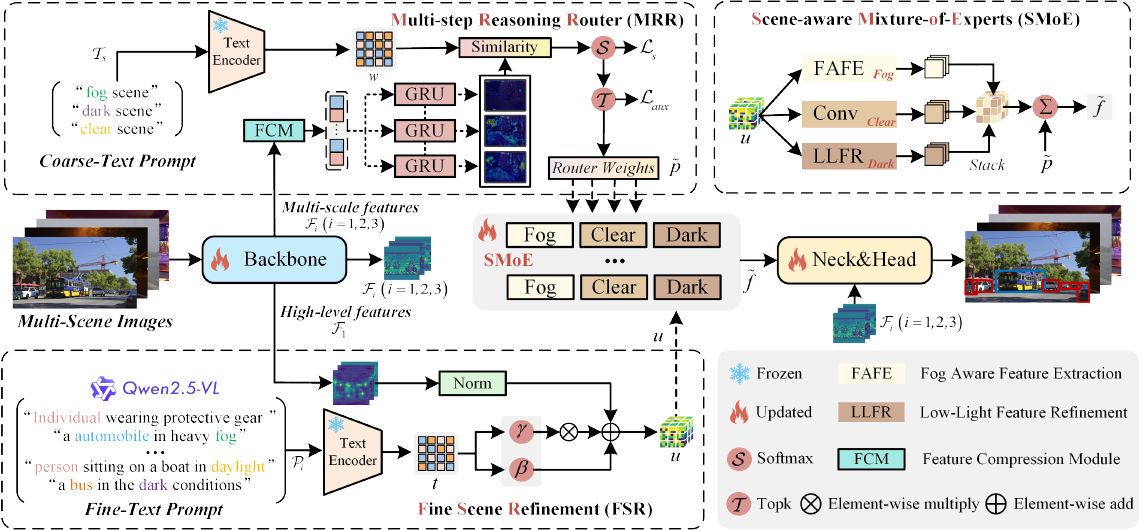}
    \caption{\textbf{Overview of proposed C2FXNet.} It consists of three key modules: the Multi-step Reasoning Router (MRR) compresses multi-scale visual cues and performs GRU-based recurrent scene reasoning under coarse text-guided prototypes to generate interpretable routing weights, the Fine Scene Refinement (FSR) refines visual features using fine-grained textual cues, and the Scene-aware Mixture-of-Experts (SMoE) dynamically fuses fog, dark, and clear experts under the joint guidance of MRR and FSR, enabling robust and scene-consistent detection. Among them, $\gamma$ and $\beta$ denote two learnable linear layers.}
    \label{fig: network}
\end{figure*}

\subsection{Vision Language Models.}
Vision–language models (VLMs) integrate visual and textual modalities into unified representations. Early architectures such as ViLBERT \cite{lu2019vilbert} and VisualBERT \cite{li2019visualbert} pioneered multimodal pretraining via transformer-based fusion. More recently, CLIP \cite{radford2021clip} and its regional extensions \cite{sun2024alphaclip, zhang2024regionword, cao2024vivl} scaled contrastive image–text learning and enhanced region–text alignment for downstream recognition and detection tasks. In parallel, large-scale open-vocabulary and LLM-based detectors \cite{xi2025owovd, fu2025llmdet, li2025benefit} further bridge visual and linguistic reasoning, demonstrating the potential of multimodal supervision for robust localization and generalization. MLLM \cite{yin2025rod} leverages large language models (LLMs) for high-level reasoning and localization, alleviating the linguistic limitations of conventional detectors. Building on these advances, our method leverages text descriptions generated by an LLM to provide global-to-local textual guidance for the Multi-step Reasoning Router and Fine Scene Refinement modules under adverse weather conditions.

\subsection{Dynamic Routing Networks.}
Dynamic routing enhances neural adaptability by conditionally activating submodules based on input features. Early work, such as Capsule Networks \cite{dynamic}, introduced routing-by-agreement to model part–whole relationships, but their limited depth constrained scalability. Later Mixture-of-Experts (MoE) models \cite{moe2017} employed learnable gating to selectively activate expert branches, improving efficiency and scalability. These ideas were extended in large-scale systems, including GShard \cite{gshard}, Switch Transformer \cite{fedus2022switch}, and DeepSeek-MoE \cite{dai2024deepseekmoe}, emphasizing routing efficiency and load balance. Recent studies further refine expert specialization for vision and large-scale networks \cite{oldfield2024multilinear,yang2024multi,zhao2025equipping}. Building upon these advances, we propose a scene-aware mixture-of-experts that adaptively routes features to scene-specific experts under adverse weather, enhancing detection robustness in diverse conditions.

\section{Method}

\noindent\textbf{Overview.} As illustrated in Figure~\ref{fig: network}, we propose a coarse-to-fine scene expert network that progressively refines scene understanding from coarse scene reasoning to local variation modeling. Given an image 
$x \! \in \! \mathbb{R}^{3 \times H \times W}$, the feature encoder 
backbone extracts multi-scale features $\{\mathcal{F}_1, \mathcal{F}_2, 
\mathcal{F}_3\}$, which are then processed by two complementary guidance modules. 
The Multi-step Reasoning Router (MRR) compresses the multi-scale features into a compact routing sequence and recurrently updates a latent routing state through GRU-based reasoning before matching the final state to frozen coarse scene prototypes (fog, dark, clear) to obtain sparse routing weights $\tilde{p} \in \mathbb{R}^3$.
In parallel, the Fine Scene Refinement (FSR) module 
uses image-specific semantic prompts to modulate the high-level feature and produce a refinement feature $u$ that captures local scene-dependent variations. The Scene-aware 
Mixture-of-Experts (SMoE) then dynamically fuses scene-specific expert 
outputs using both $\tilde{p}$ and $u$ to produce semantically aware 
representations $\tilde{f}_{\text{out}}$. These refined features are fed into a 
standard PAFPN~\cite{PAN} neck and YOLO~\cite{yolov7} detection head 
for final prediction.

The key innovation lies in coupling GRU-based coarse scene reasoning in MRR with text-guided fine feature modulation in FSR, and using their outputs to steer expert fusion in SMoE. This coarse-to-fine interaction enables robust and interpretable all-weather detection by dynamically adjusting expert contributions according to both global scene semantics and local visual variations.

\subsection{Multi-step Reasoning Router}

The proposed Multi-step Reasoning Router (MRR) performs scene-aware expert selection by combining frozen language priors with recurrent visual reasoning.
Unlike conventional routers that infer routing weights from a single visual descriptor, MRR organizes compressed multi-scale evidence into a short reasoning sequence and progressively refines the routing decision, which leads to more reliable scene assignment under adverse conditions.

\begin{figure*}
    \centering
    \includegraphics[width=1\linewidth]{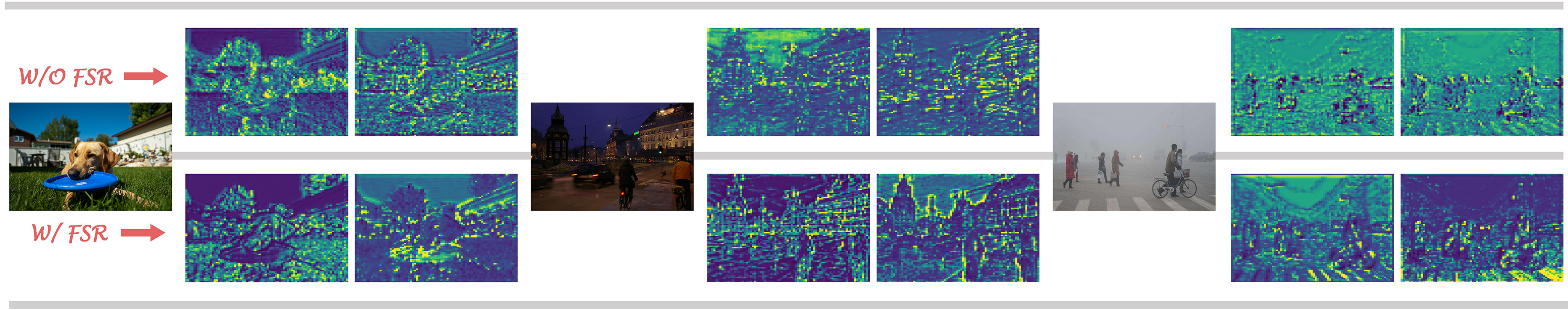}
    \caption{Feature-level interpretability of FSR in clear, dark, and foggy scenes. FSR refines noisy SMoE activations into more coherent, object-centric, and weather-consistent feature responses.}
    \label{fig:featmap}
\end{figure*}

For each coarse scene category $s \in \{\text{fog}, \text{dark}, \text{clear}\}$, we define a small set of coarse textual prompts $\mathcal{T}_s$.
A frozen text encoder $\varphi$ maps each prompt $t \in \mathcal{T}_s$ to a text embedding
\begin{equation}
    e_t = \varphi(t), \qquad e_t \in \mathbb{R}^{d_t},
\end{equation}
which is projected into the routing space by a lightweight projector $P:\mathbb{R}^{d_t}\rightarrow\mathbb{R}^{d}$.
The projected embeddings are averaged within each scene category and $\ell_2$-normalized to form a prototype $k_s \in \mathbb{R}^{d}$.
The resulting prototype matrix is
\begin{equation}
    W = [k_{\text{fog}}, k_{\text{dark}}, k_{\text{clear}}]^{\top} \in \mathbb{R}^{3 \times d}.
\end{equation}

Given multi-scale backbone features $\{\mathcal{F}_1,\mathcal{F}_2,\mathcal{F}_3\}$, the Feature Compression Module first produces an ordered set of compact routing descriptors
\begin{equation}
    \{c_n\}_{n=1}^{N} = \psi(\mathcal{F}_1,\mathcal{F}_2,\mathcal{F}_3), \qquad c_n \in \mathbb{R}^{d_h},
\end{equation}
where $\psi(\cdot)$ denotes multi-scale compression and feature re-organization, and $N$ is the number of reasoning steps.
Starting from a learnable initial state $h_0 \in \mathbb{R}^{d_h}$, MRR sequentially aggregates the routing descriptors through a GRU:
\begin{equation}
    h_n = \mathrm{GRU}(c_n, h_{n-1}), \qquad n=1,\dots,N.
\end{equation}
The final reasoning state is projected into the routing space:
\begin{equation}
    z = \Pi(h_N), \qquad z \in \mathbb{R}^{d}.
\end{equation}

MRR then computes scene affinities by matching the normalized routing feature with the fixed scene prototypes:
\begin{equation}
    p = \mathrm{softmax}\!\left(\frac{\hat{z}W^{\top}}{\tau}\right),
    \qquad
    \hat{z} = \frac{z}{\|z\|_2},
\end{equation}
where $\tau$ is a temperature parameter.
To encourage sparse expert activation, we retain only the Top-$k$ entries and renormalize the routing weights:
\begin{equation}
    \tilde{p} = \frac{p \odot m}{\sum_{s} p_s m_s + \epsilon},
\label{mrr}
\end{equation}
where $m \in \{0,1\}^{3}$ is the Top-$k$ mask.
The sparse routing weights $\tilde{p}$ are finally used to select and fuse the scene experts.
By accumulating evidence across multiple compressed cues before prototype matching, MRR yields a more stable and interpretable routing process than one-shot similarity-based assignment.

\subsection{Fine Scene Refinement}

While MRR provides coarse scene-level routing, it cannot directly enhance the adaptability of each expert to subtle visual variations. To address this limitation, the Fine Scene Refinement (FSR) module introduces fine-grained textual conditioning on the high-level backbone feature $\mathcal{F}_1$, enabling the model to focus on task-relevant regions and better align features under adverse conditions.

For each image, we utilize Qwen2.5-VL~\cite{bai2025qwen2} to generate a set of Fine-Text Prompts
\(\mathcal{P} = \{t_i\}_{i=1}^{N}\) that describe the scene and the objects in it (``an automobile in heavy fog'', ``a person sitting on a boat in daylight'').
The detailed text generation procedure is provided in the Appendix (Section A.5).
The frozen text encoder $\varphi$ maps each prompt $\mathcal{P}_i$ to an embedding $e_i = \varphi(\mathcal{P}_i)$, and the same lightweight projector $P$ used in MRR is applied to obtain projected vectors
\(z_i = P(e_i)\in\mathbb{R}^{d}\).
We then average the set \(\{z_i\}_{i=1}^{N}\) and apply $\ell_2$ normalization to obtain a single per-image textual representation
\(t \in \mathbb{R}^{d}\), which serves as the fine-text prompt.

Given the feature map $x{=}\mathcal{F}_{1}$ and the textual vector $t\in\mathbb{R}^{d}$, FSR injects fine-grained semantic information through channel-wise refinement.
We first apply Group Normalization to $x$, and then use two learnable linear layers
$\gamma(\cdot), \beta(\cdot):\mathbb{R}^{d}\!\rightarrow\!\mathbb{R}^{C}$
to generate scaling and shifting coefficients from $t$:
\begin{equation}
    \gamma(t) = W_{\gamma} t + b_{\gamma}, \qquad
    \beta(t) = W_{\beta} t + b_{\beta},
\end{equation}
where $W_{\gamma}, W_{\beta}\in\mathbb{R}^{C\times d}$ and
$b_{\gamma}, b_{\beta}\in\mathbb{R}^{C}$ are learnable parameters.
The resulting coefficients are broadcast to the spatial dimensions and applied to the normalized feature map:
\begin{equation}
    u
    =
    \mathrm{GN}(x)\cdot \big(1 + \gamma(t)\big)
    + \beta(t),
    \label{eq:fsr_u}
\end{equation}
where $\gamma(t),\beta(t)\in\mathbb{R}^{C}$ are implicitly reshaped to
$\mathbb{R}^{B\times C\times H\times W}$ for element-wise refinement.

This refinement is applied exclusively within the residual enhancement paths of each expert, leaving the backbone and MRR routing computations unchanged.
Such a decoupled design stabilizes training: MRR decides which experts to activate at the coarse scene level, whereas FSR refines how each expert adapts to local scene variations.

To analyze the role of FSR, Fig.~\ref{fig:featmap} visualizes SMoE output features with and without FSR across clear, dark, and foggy scenes. Without FSR, the activations are noisy and fragmented, showing background interference in clear scenes, illumination bias in dark scenes, and texture flattening in foggy scenes. With FSR, the responses become more concentrated on foreground objects, with clearer boundaries and less background interference. These results show that FSR complements MRR: MRR selects coarse scene experts, whereas FSR injects image-specific semantic cues to improve local discriminability after global routing.

\subsection{Scene-aware Mixture-of-Experts}
The Scene-aware Mixture-of-Experts (SMoE) module serves as the final 
adaptive stage of C2FXNet, receiving joint guidance from MRR and FSR. 
Specifically, it leverages the sparse routing weights $\tilde{p}$ from 
MRR (Eq.~\eqref{mrr}) and the scene-aware refinement feature 
$u$ from FSR (Eq.~\eqref{eq:fsr_u}) as complementary signals. 
Guided by these inputs, SMoE dynamically combines multiple scene-specific 
experts to adapt expert contributions across diverse environmental conditions.

Each expert branch specializes in one coarse scene domain (fog, dark, or clear) and operates on the refined high-level feature $u\in\mathbb{R}^{B\times C\times H\times W}$.
Within each branch, a domain-specific enhancement operator is used to capture the characteristics of its corresponding environment.
For foggy conditions, we employ a Feature-Aware Frequency Enhancement (FAFE) operator to restore high-frequency details and contrast:
\begin{equation}
    \mathrm{FAFE}(u)
    = \mathrm{LeakyReLU}\big((u' \odot u) - u' + 1\big),
\end{equation}
where $u'{=}\mathrm{Conv}_{1\times1}(u)$ is a linear projection and $\odot$ denotes element-wise multiplication.
For low-light scenes, the dark expert applies a Low-Light Frequency Restoration (LLFR) operator to rebalance illumination:
\begin{equation}
    \mathrm{LLFR}(u)
    = \mathrm{LeakyReLU}\!\left((\alpha|u' + u| + \epsilon)^{\rho}\right),
\end{equation}
where $\alpha$, $\rho$ and $\epsilon$ are learnable or fixed hyperparameters controlling the enhancement strength.
The clear expert uses two standard convolutional layers to provide neutral enhancement in well-lit conditions.
Let $v_s^{(g)}\in\mathbb{R}^{B\times C\times H\times W}$ denote the output of scene expert $s\in\{\text{fog},\text{dark},\text{clear}\}$ in the $g$-th expert group when processing $u$.
Given the routing weights $\tilde{p}=[\tilde{p}_{\text{fog}},\tilde{p}_{\text{dark}},\tilde{p}_{\text{clear}}]$ from MRR, the fused output of the $g$-th group is
\begin{equation}
    \tilde{f}^{(g)} = \sum_{s} \tilde{p}_s\, v_s^{(g)}, \qquad
    \tilde{f}_{\text{out}} = \frac{1}{G}\sum_{g=1}^{G}\tilde{f}^{(g)},
\end{equation}
where $G$ is the total number of expert groups.
In practice, $G$ controls the model capacity, while the same sparse routing weights $\tilde{p}$, inferred by MRR through recurrent scene reasoning, are shared across groups. 
The fused feature $\tilde{f}_{\text{out}}$ is both scene-consistent and context-aware, providing a unified representation for robust object detection under adverse weather and lighting conditions.

\begin{figure}
    \centering
    \includegraphics[width=1\linewidth]{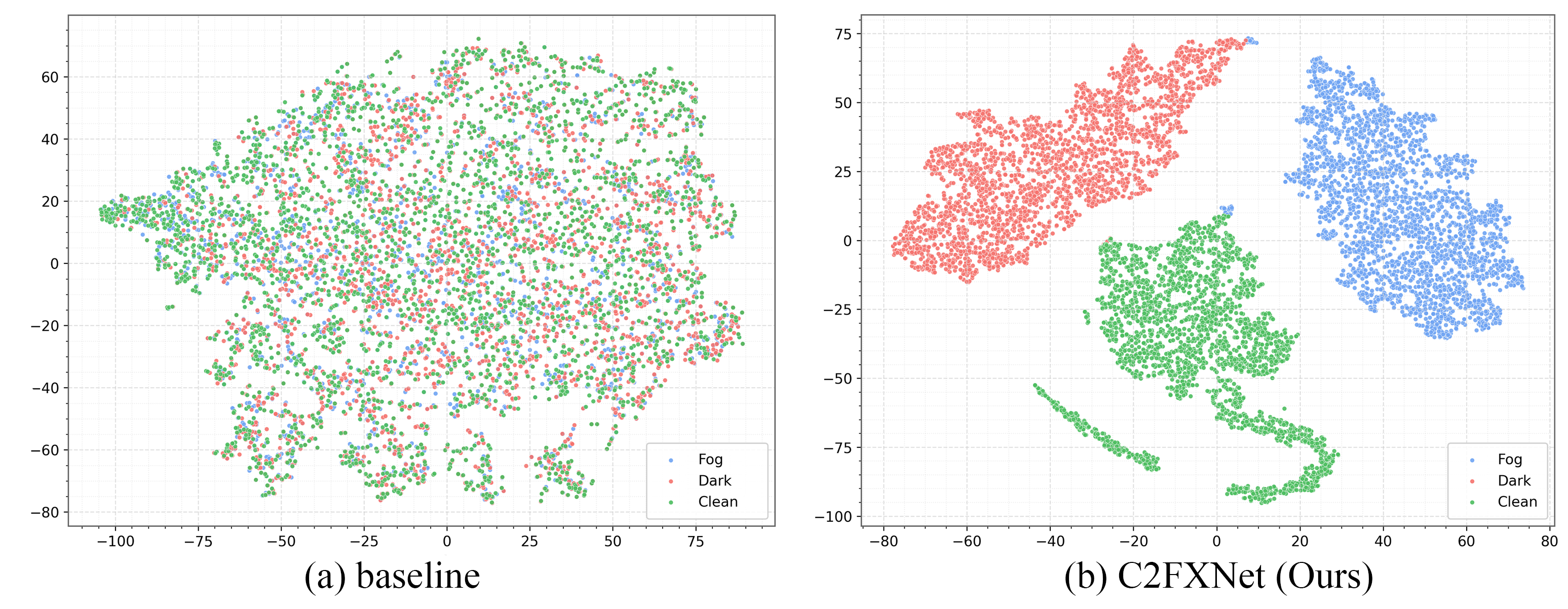}
    \caption{t-SNE visualization of feature embeddings. C2FXNet produces more compact and better-separated clusters than the baseline for fog, dark, and clear scenes.}
    \label{fig:tsne}
\end{figure}

As visualized by the t-SNE embeddings in Fig.~\ref{fig:tsne}, the baseline features exhibit severe overlap across fog/dark/clear domains, indicating weak scene discriminability. In contrast, C2FXNet produces clearly separated clusters with improved intra-domain compactness, validating that SMoE yields a more scene-consistent representation. By coupling the coarse recurrent reasoning of MRR with the fine-grained semantic modulation of FSR, SMoE bridges global semantic understanding and local feature adaptation. The fused feature $\tilde{f}_{\text{out}}$ is therefore both scene-consistent and context-aware, providing a unified representation for robust object detection under adverse weather and lighting conditions.
Finally, C2FXNet is trained end-to-end with a joint loss that combines detection, scene classification, and a
load-balancing loss on the MRR:
\begin{equation}
    \mathcal{L}_{total}
    = \mathcal{L}_{det}
    + \lambda_1 \mathcal{L}_{\text{s}}
    + \lambda_2\mathcal{L}_{aux},
\end{equation}
where $\mathcal{L}_{det}$ is the standard YOLO detection loss \cite{yolov7}, and
$\mathcal{L}_s$ is a cross-entropy loss supervising the scene prediction branch.
The MRR load-balancing loss $\mathcal{L}_{aux}$ is computed from the
routing weights over $R$ experts:
\begin{equation}
    \mathcal{L}_{aux}
    = R \sum_{r=1}^{R} \bar{r}_r^{\,2},
    \qquad
    \bar{r} = \frac{1}{N}\sum_{i=1}^{N}r_i,
\end{equation}
where $r_i \in \mathbb{R}^3$ denotes the MRR routing weights for the $i$-th image.
This term encourages balanced expert utilization within the MRR.

\section{Experiments}
\label{sec: experiments}

\paragraph{Datasets and Evaluation Metrics.} 
We evaluate our method on three datasets: Adverse Weather Dataset (AWD), RTTS~\cite{RTTS-li2018benchmarking}, and ExDark~\cite{ExDark-loh2019getting}. 
AWD is a synthetic dataset we constructed using CycleGAN~\cite{cyclegan} applied to MS COCO~\cite{lin2014microsoft} to simulate foggy and low-light conditions for both training and evaluation. 
RTTS and ExDark contain real foggy and low-light images, respectively, while Rain and Snow are synthetic datasets created by applying physical degradation models to VOC images. 
Table~\ref{tab1} summarizes the statistics and label distributions of each dataset.

To ensure fair comparison, all datasets share five common object categories. 
We use Parameters (Params), Floating Point Operations 
(FLOPs), and mean Average Precision (mAP) for evaluation.

\paragraph{Implementation Details.} The Stochastic Gradient Descent (SGD) is applied as the optimizer with an initial learning rate of 0.01. The momentum is set to 0.937, and weight decay is set to 5e-4. The batch size is set to 16. We utilize NVIDIA GeForce RTX3090 GPU to train the model for 100 epochs. 
For hyperparameter settings, we empirically set the temperature coefficient $\tau{=}1$ in MRR, Top-$k{=}1$ for sparse expert selection, and $\epsilon{=}10^{-9}$ for normalization stability. In the LLFR operator, $\alpha{=}0.5$, $\rho{=}1.2$, and $\epsilon{=}10^{-6}$ are adopted for illumination restoration. For the overall loss, the weighting coefficients are $\lambda_1{=}0.1$ and $\lambda_2{=}0.05$ to balance scene classification and routing regularization.
To obtain fine-grained textual descriptions for instance-level refinement, we employ Qwen2.5-VL~\cite{bai2025qwen2} deployed via the Ollama framework, which generates multiple object-centric textual descriptions for each image.
Additional experimental settings are provided in the Appendix (Section~A.8).

\begin{table}[!t]
\centering
\caption{Statistics of datasets used for training and evaluation.}
\label{tab1}
\setlength{\tabcolsep}{3pt}
\resizebox{\linewidth}{!}{
\begin{tabular}{lcccccccc}
\toprule
\multirow{2}{*}{\textbf{Dataset}} & 
\multirow{2}{*}{\textbf{Scene}} & 
\multirow{2}{*}{\textbf{\#Img}} &
\multicolumn{5}{c}{\textbf{\#Bounding Boxes / Class}} \\
\cmidrule(lr){4-8}
 &  &  & \textbf{Pers.} & \textbf{Bic.} & \textbf{Car} & \textbf{Moto.} & \textbf{Bus} \\
\midrule
AWD (Train) & Fog/Dark/Clear & 41{,}685 & 157{,}707 & 4{,}014 & 26{,}064 & 4{,}875 & 3{,}489 \\
AWD (Test)  & Fog/Dark/Clear & 8{,}805  & 33{,}012  & 948   & 5{,}796 & 1{,}113 & 855 \\
RTTS         & Fog            & 4{,}322  & 7{,}950   & 534   & 18{,}415 & 862 & 1{,}838 \\
ExDark    & Dark           & 7{,}363  & 7{,}460   & 1{,}120 & 2{,}927 & 1{,}072 & 706 \\
\bottomrule
\end{tabular}
}
\end{table}

\begin{table*}[!t]
\small
\renewcommand{\arraystretch}{0.9}
  \belowrulesep=0.2pt
  \aboverulesep=0.2pt
  \centering
    \caption{Comparison of C2FXNet with previous detectors on RTTS, ExDark, and AWD datasets. C2FXNet attains optimal parameter efficiency while achieving the highest detection accuracy. The best and second-best results are boldfaced and underlined.}
  \setlength{\tabcolsep}{1.4mm}
   \begin{tabularx}{\textwidth}{>{\centering\arraybackslash}p{3.1cm}|>{\centering\arraybackslash}p{1.3cm}|>{\centering\arraybackslash}p{2.3cm}|>{\centering\arraybackslash}p{1.3cm}|>{\centering\arraybackslash}p{1.3cm}|c|c|c|c|c|c}
    \toprule
    Method  & Dataset & Publication & Params $\downarrow$ & FLOPs $\downarrow$ & Bicycle & Bus   & Car   & Motorbike & Person & mAP (\%) $\uparrow$ \\
    \midrule
    TogetherNet \cite{togethernet-wang2022} & \multirow{9}{*}{RTTS} & \textit{CGF'22} & 15.7M & 26.1G & 49.50  & 27.06  & 70.86  & 46.69  & 81.66  & 55.15  \\
    IA-YOLO \cite{iayolo-liu2022image}&       & \textit{AAAI'22} & 58.9M & 65.6G & 39.06  & 17.87  & 41.29  & 30.63  & 58.76  & 37.52  \\
    GDIP-YOLO \cite{gdip-kalwar2023}&       & \textit{ICRA'23} & 68.0M & 105.2G & 45.89  & 11.58  & 49.50  & 32.96  & 71.98  & 42.38  \\
    YOLOv9s \cite{wang2024yolov9}&       & \textit{ECCV'24} & 7.1M  & 26.4G & 40.73  & 23.41  & 61.13  & 33.17  & 79.76  & 47.64  \\
    YOLOv10s \cite{wang2024yolov10}&       & \textit{NeurIPS'24} & 7.6M  & 24.5G & 42.22  & 23.65  & 60.42  & 33.60  & 78.72  & 47.72  \\
    YOLOv11s \cite{yolov11}&       & \textit{-} & 9.0M  & 21.5G & 53.64  & 37.68  & 72.56  & \underline{53.23}  & 82.87  & 60.00  \\
    YOLOv12s \cite{tian2025yolov12}&       & \textit{NeurIPS'25} & 8.6M  & \underline{19.3G} & \underline{58.30}  & 31.01 & 69.67 & 52.64 & 81.43 & 58.61 \\
    YOLOv13s \cite{lei2025yolov13}&       & \textit{Arxiv'25} & 8.6M  & 21.7G & 54.75  & 39.15  & 72.80  & 52.61  & \underline{83.77}  & 60.62  \\
    EEnvA-Mamba \cite{chen2026eenva}&       & \textit{PR'26} & \underline{6.9M}  & \textbf{15.6G} & 53.80  & \textbf{71.60}  & \textbf{77.60}  & 47.00  & 53.40  & \underline{60.70}  \\
    \rowcolor{gray!12}
    \textbf{C2FXNet (Ours)} &       & \textit{-} & \textbf{6.7M}  & 22.0G & \textbf{57.90}  & \underline{41.90}  & \underline{76.53}  & \textbf{57.09}  & \textbf{85.08}  & \textbf{63.70}  \\
    \midrule\midrule
    Zero-DCE \cite{guo2020zero}& \multirow{9}{*}{ExDark} & \textit{CVPR'20} & 12.1M & 13.9G & 50.21  & 61.21  & 46.00  & 37.35  & 49.72  & 48.90  \\
    IA-YOLO \cite{iayolo-liu2022image}&       & \textit{AAAI'22} & 58.9M & 65.6G & \underline{68.47}  & 72.56  & 59.96  & 53.77  & \underline{73.31}  & 65.61  \\
    GDIP-YOLO \cite{gdip-kalwar2023}&       & \textit{ICRA'23} & 68.0M & 105.2G & 53.31  & 62.79  & 42.72  & 39.09  & 49.72  & 49.53  \\
    YOLOv9s \cite{wang2024yolov9}&       & \textit{ECCV'24} & \underline{7.1M}  & 26.4G & 51.35  & 66.61  & 46.97  & 43.67  & 60.56  & 53.83  \\
    YOLOv10s \cite{wang2024yolov10}&       & \textit{NeurIPS'24} & 7.6M  & 24.5G & 53.47  & 61.69  & 45.65  & 46.52  & 59.22  & 53.31  \\
    YOLOv11s \cite{yolov11}&       & \textit{-} & 9.0M  & 21.5G & 64.36  & \textbf{80.87}  & \underline{62.82}  & 60.80  & 70.19  & \underline{67.81}  \\
    YOLOv12s \cite{tian2025yolov12}&       & \textit{NeurIPS'25} & 8.6M  & \textbf{19.3G} & 64.62 & 76.31 & 60.32 & \underline{61.75} & 68.42 & 66.23 \\
    YOLOv13s \cite{lei2025yolov13}&       & \textit{Arxiv'25} & 8.6M  & 21.7G & 65.53  & 77.92  & 59.40  & 60.15  & 71.28  & 66.86  \\
    \rowcolor{gray!12}
    \textbf{C2FXNet (Ours)} &       & \textit{-} & \textbf{6.7M}  & \underline{22.0G} & \textbf{68.65}  & \underline{81.16}  & \textbf{66.59}  & \textbf{63.24}  & \textbf{76.08}  & \textbf{71.14}  \\
    \midrule\midrule
    YOLOv9s \cite{wang2024yolov9}& \multirow{6}{*}{AWD} & \textit{ECCV'24} & \underline{7.1M}  & 26.4G & 22.21  & 54.20  & 38.27  & 41.20  & 60.28  & 43.23  \\
    YOLOv10s \cite{wang2024yolov10}&       & \textit{NeurIPS'24} & 7.6M  & 24.5G & 22.80  & 53.99  & 37.31  & 43.49  & 59.53  & 43.42  \\
    YOLOv11s \cite{yolov11}&       & \textit{-} & 9.0M  & 21.5G & 33.72  & \textbf{65.32}  & \textbf{49.15}  & 54.06  & \underline{67.08}  & \underline{53.87}  \\
    YOLOv12s \cite{tian2025yolov12}&       & \textit{NeurIPS'25} & 8.6M  & \textbf{19.3G} & \underline{34.51} & 64.23 & 46.87 & \textbf{56.30}  & 66.93 & 53.77 \\
    YOLOv13s \cite{lei2025yolov13}&       & \textit{Arxiv'25} & 8.6M  & 21.7G & 33.82  & 62.63  & 45.70  & 53.55  & 63.17  & 51.17  \\
    \rowcolor{gray!12}
    \textbf{C2FXNet (Ours)} &       & \textit{-} & \textbf{6.7M}  & \underline{22.0G} & \textbf{34.73}  & \underline{64.25}  & \underline{48.02}  & \underline{54.45}  & \textbf{69.50} & \textbf{54.19}  \\ 
    \bottomrule
   \end{tabularx}%
  \label{tab2}%
\end{table*}

\begin{figure*}[!t]
    \centering
    \includegraphics[width=1\linewidth]{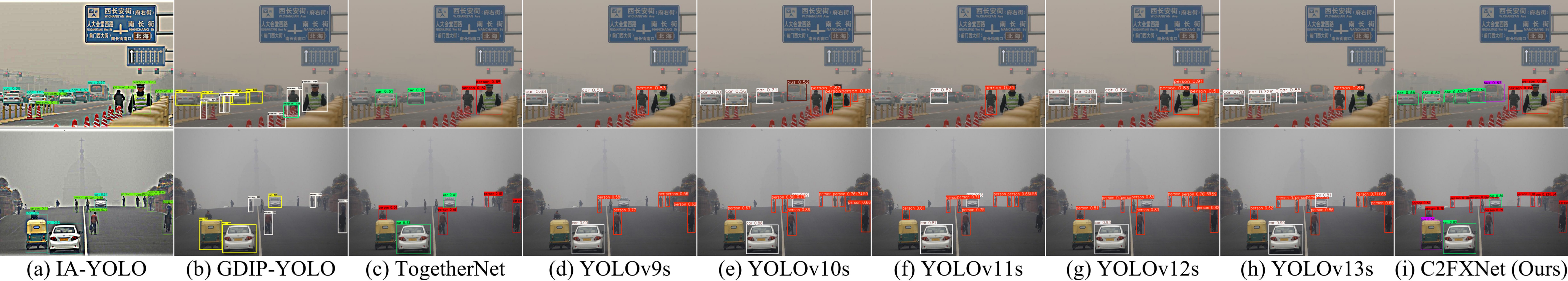}
    \caption{Visual comparison on the RTTS dataset under foggy conditions. C2FXNet detects more objects with clearer boundaries and higher confidence, especially for small or distant targets that other methods miss.}
    \label{fig:rtts}
\end{figure*}

\subsection{Comparisons with State-of-the-Art Methods}

We compare C2FXNet with state-of-the-art detectors, including TogetherNet~\cite{togethernet-wang2022}, IA-YOLO~\cite{iayolo-liu2022image}, GDIP-YOLO~\cite{gdip-kalwar2023}, and recent YOLO variants from YOLOv9s to YOLOv13s~\cite{wang2024yolov9,wang2024yolov10,yolov11,tian2025yolov12,lei2025yolov13} and EEnvA-Mamba~\cite{chen2026eenva}. All methods are evaluated on RTTS, ExDark, and AWD, covering foggy, low-light, and clear-weather conditions.

\subsubsection{Quantitative Results on All-Weather Benchmarks}
Table~\ref{tab2} presents comprehensive comparisons across three benchmarks. 
On RTTS, C2FXNet achieves 63.70\% mAP, surpassing YOLOv13s by 3.08\% with notable gains in Bus (+2.75\%), Car (+3.73\%), Motorbike (+4.48\%), and Person (+1.31\%). 
These improvements indicate that our semantic-guided feature refinement effectively preserves discriminative object structures under fog-induced degradation, especially for categories sensitive to blurred boundaries and low-contrast appearance. 

On ExDark, our method reaches 71.14\% mAP, outperforming YOLOv11s by 3.33\% and Zero-DCE by 22.24\%. 
The substantial margin over enhancement-based Zero-DCE highlights the advantage of performing feature-space adaptation with semantic priors, which avoids pixel-level artifacts while strengthening missing scene cues under severe low-light conditions. 
Meanwhile, compared with recent lightweight YOLO variants, C2FXNet consistently improves all categories and achieves the best overall accuracy, demonstrating the effectiveness of the proposed coarse-to-fine expert routing strategy for low-light perception.

On AWD, C2FXNet achieves 54.19\% mAP, exceeding YOLOv11s and YOLOv12s by 0.32\% and 0.42\%, respectively. 
Although YOLOv11s performs slightly better on Bus and Car, our model achieves higher AP on Bicycle and Person and yields the best overall mAP. 
This suggests that C2FXNet favors more balanced cross-scene generalization rather than over-optimizing a few dominant categories. 
In particular, the consistent gains on challenging classes such as Motorbike and Person across RTTS and ExDark further verify the robustness of the proposed recurrent scene-aware expert routing mechanism under diverse adverse weather conditions.

\begin{figure*}[!t]
    \centering
    \includegraphics[width=1\linewidth]{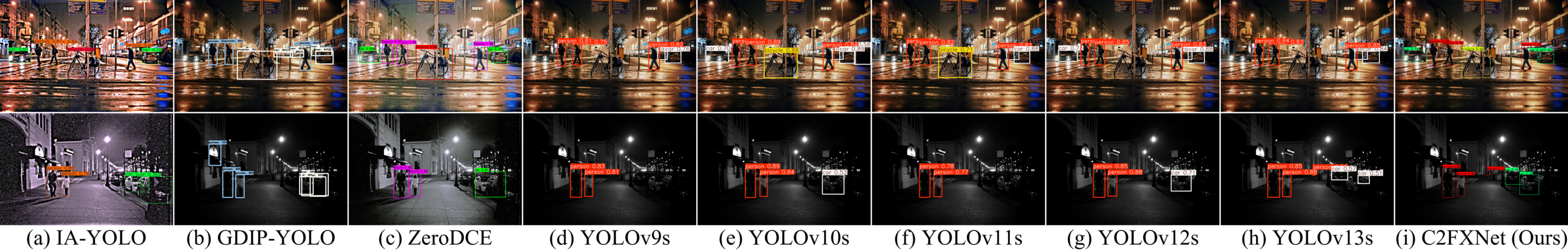}
    \caption{Visual comparison on the ExDark dataset under low-light conditions. C2FXNet achieves more accurate and confident detections, successfully identifying objects in dark and overexposed areas.}
    \label{fig:exdark}
\end{figure*}

\textbf{Scene-wise Adaptation Analysis on AWD.} To validate the scene-adaptive capability of our text-guided framework, we compare C2FXNet with the baseline (YOLOv7-T \cite{yolov7}) across different weather conditions on the AWD dataset. Table~\ref{tab3} presents the per-scene breakdowns. 
In foggy scenes, C2FXNet achieves 53.28\% mAP (+1.19\%), with notable 
gains in Bus (+2.24\%) and Motorbike (+2.36\%). This improvement indicates 
that MRR can accumulate reliable coarse scene evidence before dispatching 
features to the fog expert, while FSR further strengthens object-aware 
responses under haze-induced contrast degradation. In dark scenes, 
C2FXNet reaches 51.30\% mAP (+1.26\%), with the largest improvement on 
Motorbike (+3.79\%). This gain validates the complementarity between MRR 
and FSR: the former stabilizes expert selection under severe illumination 
ambiguity, whereas the latter selectively amplifies object-relevant channels 
from semantic descriptions such as ``a motorcycle with dim headlights''. 
Under clear weather, C2FXNet attains 59.22\% mAP (+0.77\%), with strong 
gains in Bus (+3.10\%) and Motorbike (+2.26\%). These improvements show 
that the proposed semantic priors remain useful even in non-degraded scenes, 
where coarse scene reasoning and fine textual cues help resolve spatial 
ambiguities in crowded layouts.
\begin{figure}[!t]
    \centering
    \includegraphics[width=1\linewidth]{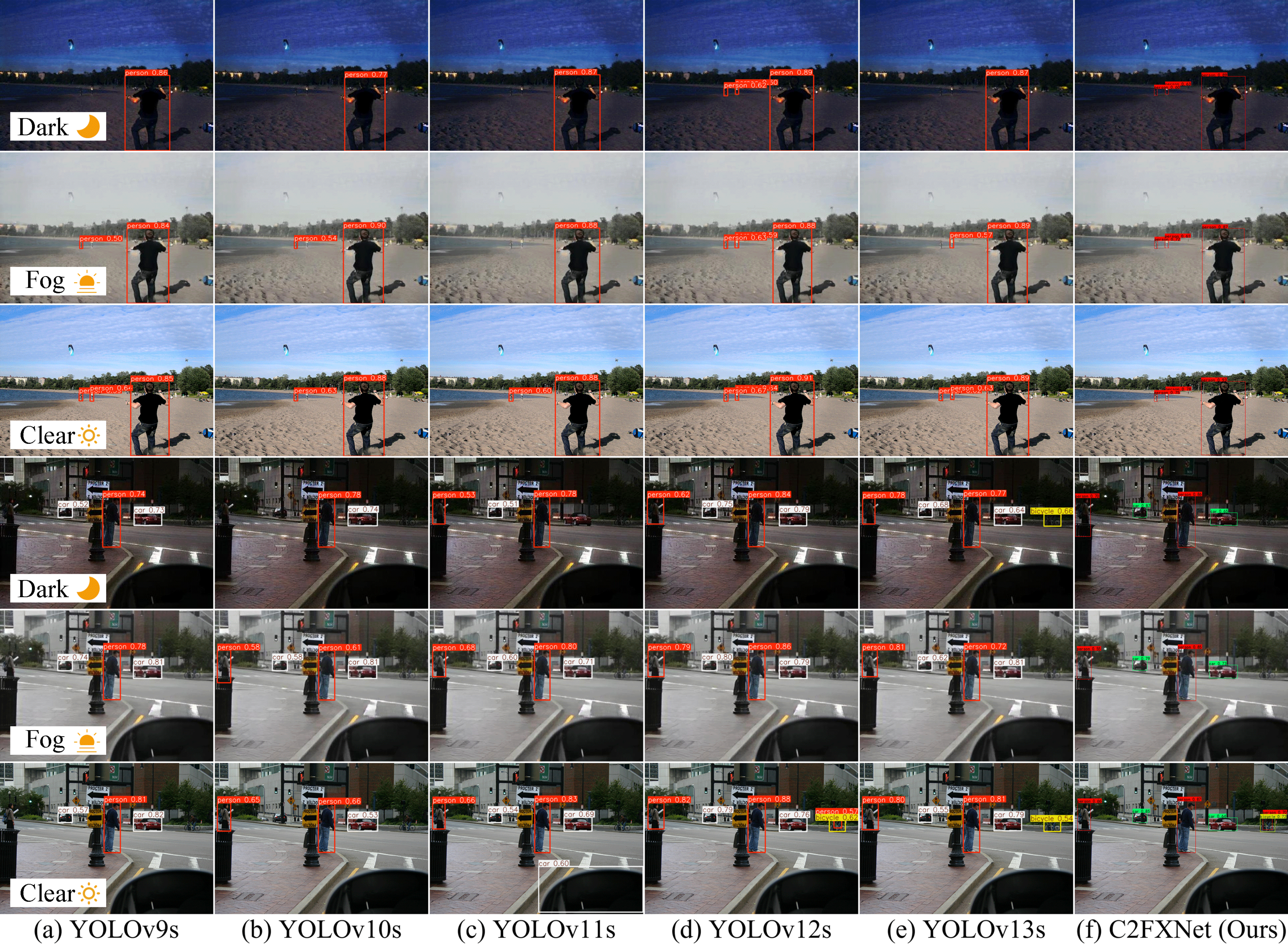}
    \caption{Visual comparison on the AWD dataset under dark, foggy, and clear conditions. C2FXNet maintains consistent detection quality and higher confidence across scenes, highlighting its strong cross-weather adaptability.}
    \label{fig:awd}
\end{figure}

\subsubsection{Qualitative Results on All-Weather Scenes}
To further substantiate the quantitative, we present qualitative comparisons between C2FXNet and the representative approach on the RTTS, ExDark, and AWD datasets, covering diverse weather conditions. 

\textbf{Foggy Scenes (RTTS).}
As shown in Figure~\ref{fig:rtts}, C2FXNet yields denser and more accurate detections in heavy fog. IA-YOLO, GDIP-YOLO, and recent YOLO variants often miss distant pedestrians or partially occluded vehicles because fog severely suppresses edges and contrast. With MRR progressively reasoning over compressed multi-scale cues before routing features toward fog experts, together with FAFE-based enhancement, our model better restores high-frequency structures and keeps small or distant objects separable from the hazy background.

\begin{table}[!t]
\belowrulesep=0.2pt
\aboverulesep=0.2pt
\centering
\caption{Comparison of C2FXNet and the baseline on the AWD dataset under fog, dark, and clear scenes. C2FXNet yields consistent gains across most categories, highlighting the effectiveness of scene-aware adaptation.}
\label{tab3}
\resizebox{\columnwidth}{!}{
\begin{tabular}{l|c|c|c|c|c|c|c}
\toprule
Method & Scene & Bicycle & Bus & Car & Motorbike & Person & mAP (\%) $\uparrow$\\
\midrule
Baseline      & \multirow{2}{*}{Fog}   & 33.07 & 61.34 & \textbf{45.77} & 54.58 & 65.67 & 52.09 \\
C2FXNet (Ours)&                        & \textbf{34.46} & \textbf{63.58} & 45.01 & \textbf{56.94} & \textbf{66.42} & \textbf{53.28} \\
\midrule\midrule
Baseline      & \multirow{2}{*}{Dark}  & 30.48 & 62.62 & \textbf{44.61} & 49.38 & 63.11 & 50.04 \\
C2FXNet (Ours)&                        & \textbf{31.66} & \textbf{63.91} & 44.13 & \textbf{53.17} & \textbf{63.61} & \textbf{51.30} \\
\midrule\midrule
Baseline      & \multirow{2}{*}{Clear} & 41.79 & 65.47 & \textbf{54.63} & 57.88 & 72.48 & 58.45 \\
C2FXNet (Ours)&                        & \textbf{42.61} & \textbf{68.57} & 54.58 & \textbf{60.14} & \textbf{73.77} & \textbf{59.22} \\
\bottomrule
\end{tabular}
}
\end{table}

\begin{table*}[!t]
\centering
\scriptsize
\belowrulesep=0.1pt
\aboverulesep=0.1pt
\caption{Ablation for C2FXNet components on all-weather benchmarks. We adopt two frozen text encoders: \textit{all-MiniLM-L6-v2} and \textit{ViT-B-32}. “$n$ × SMoE” denotes the number of expert groups, each containing three sub-experts corresponding to the above scenes, while multiple groups are fused by different aggregation strategies (\textit{mean}, \textit{sum}, \textit{concat}). “\cmark” and “\xmark” denote with and without each component, respectively.}
\label{tab4}
\resizebox{\linewidth}{!}{
\begin{tabular}{>{\centering\arraybackslash}p{1cm}|cc>{\centering\arraybackslash}p{2.5cm}cc|c|c|c}
\toprule
\multicolumn{1}{c|}{\multirow{2}{*}{Ablation}} 
& \multirow{2}{*}{MRR} 
& \multirow{2}{*}{FSR} 
& \multirow{2}{*}{Text Encoder} 
& \multirow{2}{*}{SMoE} 
& \multirow{2}{*}{Aggregation} 
& \multicolumn{3}{c}{mAP (\%) $\uparrow$} \\
\cmidrule{7-9}
\multicolumn{1}{c|}{} &  &  &  &  &  & RTTS & ExDark & AWD \\
\midrule
(a) & \xmark & \xmark & \xmark & \xmark & – 
    & 61.99 & 69.16 & 53.53 \\
(b) & \cmark & \xmark & \xmark & $1\times$ & – 
    & 58.91$_{\text{\textcolor[rgb]{0,0.7,0}{-3.08}}}$ 
    & 69.74$_{\text{\textcolor[rgb]{0.8,0,0}{+0.58}}}$ 
    & 48.61$_{\text{\textcolor[rgb]{0,0.7,0}{-4.92}}}$ \\
(c) & \xmark & \cmark & \xmark & $1\times$ & – 
    & 60.05$_{\text{\textcolor[rgb]{0,0.7,0}{-1.94}}}$ 
    & 69.57$_{\text{\textcolor[rgb]{0.8,0,0}{+0.41}}}$ 
    & 48.80$_{\text{\textcolor[rgb]{0,0.7,0}{-4.73}}}$ \\
(d) & \cmark & \xmark & \textit{all-MiniLM-L6-v2} & $1\times$ & – 
    & 61.38$_{\text{\textcolor[rgb]{0,0.7,0}{-0.61}}}$ 
    & 69.95$_{\text{\textcolor[rgb]{0.8,0,0}{+0.79}}}$ 
    & 52.09$_{\text{\textcolor[rgb]{0,0.7,0}{-1.44}}}$ \\
(e) & \cmark & \cmark & \textit{ViT-B-32} & $1\times$ & – 
    & 59.72$_{\text{\textcolor[rgb]{0,0.7,0}{-2.27}}}$ 
    & 71.03$_{\text{\textcolor[rgb]{0.8,0,0}{+1.87}}}$ 
    & 50.16$_{\text{\textcolor[rgb]{0,0.7,0}{-3.37}}}$ \\
(f) & \cmark & \cmark & \textit{all-MiniLM-L6-v2} & $2\times$ & \textit{Mean} 
    & 59.21$_{\text{\textcolor[rgb]{0,0.7,0}{-2.78}}}$ 
    & 70.18$_{\text{\textcolor[rgb]{0.8,0,0}{+1.02}}}$ 
    & 50.58$_{\text{\textcolor[rgb]{0,0.7,0}{-2.95}}}$ \\
(g) & \cmark & \cmark & \textit{all-MiniLM-L6-v2} & $2\times$ & \textit{Sum} 
    & 58.88$_{\text{\textcolor[rgb]{0,0.7,0}{-3.11}}}$ 
    & 71.29$_{\text{\textcolor[rgb]{0.8,0,0}{+2.13}}}$ 
    & 49.83$_{\text{\textcolor[rgb]{0,0.7,0}{-3.70}}}$ \\
(h) & \cmark & \cmark & \textit{all-MiniLM-L6-v2} & $2\times$ & \textit{Concat} 
    & 61.12$_{\text{\textcolor[rgb]{0,0.7,0}{-0.87}}}$ 
    & \textbf{71.98$_{\text{\textcolor[rgb]{0.8,0,0}{+2.82}}}$}
    & 49.91$_{\text{\textcolor[rgb]{0,0.7,0}{-3.62}}}$ \\
(i) & \cmark & \cmark & \textit{all-MiniLM-L6-v2} & $1\times$ & – 
    & \textbf{63.70$_{\text{\textcolor[rgb]{0.8,0,0}{+1.71}}}$} 
    & 71.14$_{\text{\textcolor[rgb]{0.8,0,0}{+1.98}}}$ 
    & \textbf{54.19$_{\text{\textcolor[rgb]{0.8,0,0}{+0.66}}}$} \\
\bottomrule
\end{tabular}
}
\end{table*}

\textbf{Low-Light Scenes (ExDark).}
Figure~\ref{fig:exdark} presents detection results on nighttime scenes with extreme illumination variations. Enhancement-based approaches like Zero-DCE tend to introduce ringing artifacts and noise, which translate into false positives and inaccurate box boundaries, while the YOLO series frequently misses small or distant objects that are barely visible against dark backgrounds. By incorporating the FSR module, C2FXNet injects fine-text prompt (``vehicle in dark street'' or ``person near bright sign'') into the high-level features and applies the LLFR operator to selectively amplify informative channels while suppressing illumination noise. As a result, our model produces more stable confidence scores, reduces background clutter, and yields more precise localizations in underexposed regions than both enhancement-based and purely convolutional methods.

\textbf{Multi-Scenes (AWD).}
As shown in Figure~\ref{fig:awd}. In clear weather, C2FXNet and strong YOLO variants such as YOLOv12s generate similarly accurate boxes on large, well-illuminated objects. Under adverse conditions, however, the advantage of our design becomes more pronounced: in dark and foggy scenes with mixed degradations and spatially varying visibility, only C2FXNet can stably detect most objects across all categories, while YOLOv9s--YOLOv11s produce many false negatives and fragmented boxes, especially for small or partially occluded targets. This robustness stems from the joint effect of MRR, FSR, and SMoE: MRR recurrently accumulates coarse scene evidence before routing, the fine-text prompts adapt feature refinement to the current image, and the expert fusion mechanism balances their contributions across scene-specific branches.

\subsection{Ablation Studies}
\begin{figure}[!t]
    \centering
    \includegraphics[width=1\linewidth]{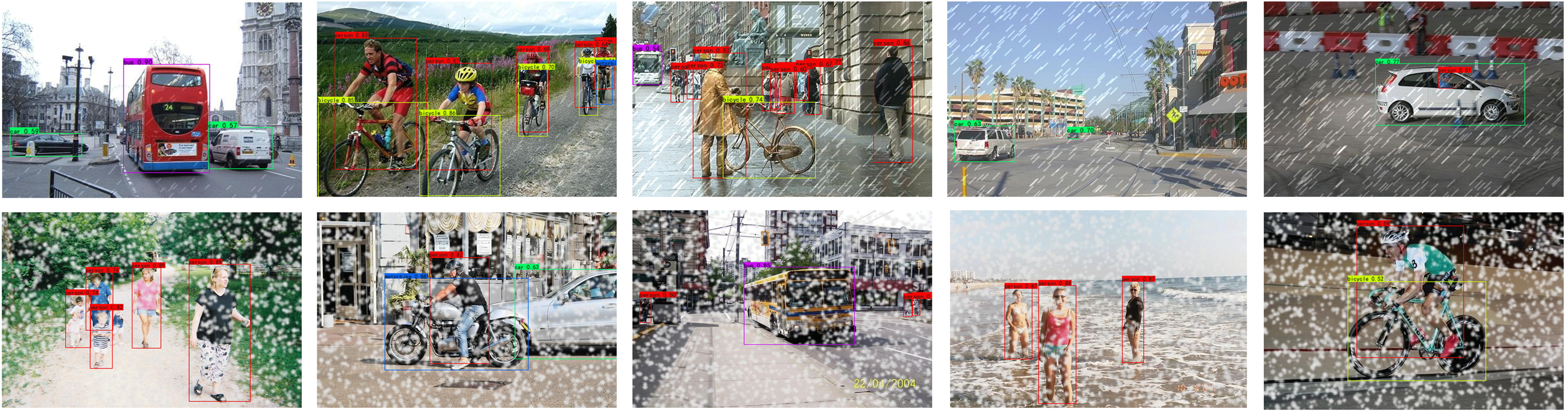}
    \caption{Visualization of detection results produced by C2FXNet on the  VOC-Rain and VOC-Snow datasets.}
    \label{fig:snow_rain}
\end{figure}

As shown in Table~\ref{tab4}, our ablation study shows that C2FXNet benefits from the collaboration of MRR, FSR, the frozen text encoder, and SMoE, rather than any single module. MRR performs coarse recurrent scene reasoning, while FSR provides semantically aligned local modulation, jointly enabling reliable expert routing and adaptive fusion.

From rows (b)--(d), enabling only partial routing-related components yields limited gains and often degrades performance. For example, row (b) drops by $3.08\%$ on RTTS and $4.92\%$ on AWD compared with the baseline, indicating that expert fusion alone cannot provide stable routing without sufficient semantic guidance. After introducing the frozen text encoder in row (d), performance becomes notably more stable, especially on AWD, which is reduced to only $1.44\%$ below the baseline, validating the benefit of coarse language priors for scene discrimination.
Rows (e)--(h) further evaluate the effects of FSR, text encoder choice, and multiple expert groups. With FSR enabled, the model consistently improves on ExDark, e.g., achieving gains of $+1.87\%$ in row (e) and $+2.82\%$ in row (h), which demonstrates the advantage of semantic alignment under low-light conditions. However, increasing the number of expert groups or changing the aggregation strategy leads to mixed results suggesting that simply enlarging expert capacity does not necessarily improve cross-domain robustness.

When all components are jointly activated in row (i), C2FXNet achieves the best overall trade-off and delivers consistent improvements across all benchmarks, with gains of $+1.71\%$ on RTTS, $+1.98\%$ on ExDark, and $+0.66\%$ on AWD over the baseline. This verifies that the full model forms an effective cooperative framework that unifies recurrent scene reasoning, text-aware local modulation, and adaptive expert fusion for robust all-weather object detection.

\begin{table}[!t]
  \centering
  \belowrulesep=0.2pt
  \aboverulesep=0.2pt
  \caption{Quantitative evaluation on VOC-Rain and VOC-Snow, comparing the baseline model with our C2FXNet across five object categories and the overall mAP.}
  \resizebox{\columnwidth}{!}{
  \begin{tabular}{c|c|c|c|c|c|c|c}
    \toprule
    Method & Dataset & Bicycle & Bus & Car & Motorbike & Person & mAP $(\%) \uparrow$ \\
    \midrule
    Baseline & \multirow{2}{*}{Rain}
             & \textbf{40.94} & 67.57 & \textbf{50.11} & \textbf{46.72} & 65.26 & 54.12 \\
    C2FXNet (Ours) &
             & 37.65 & \textbf{74.96} & 48.82 & 46.40 & \textbf{68.22} & \textbf{55.21} \\
    \midrule
    Baseline & \multirow{2}{*}{Snow}
             & \textbf{31.33} & 59.27 & 30.18 & 30.81 & 52.32 & 40.78 \\
    C2FXNet (Ours) &
             & 28.28 & \textbf{61.42} & \textbf{33.50} & \textbf{30.90} & \textbf{55.85} & \textbf{41.99} \\
    \bottomrule
  \end{tabular}}
  \label{tab:voc-weather}
\end{table}

\textbf{Generalization of rain and snow weather.}
To further evaluate the robustness and cross-weather generalization, we evaluate C2FXNet on two challenging adverse-weather benchmarks: VOC-Rain and VOC-Snow. Both datasets are constructed by synthesizing rain and snow degradations on PASCAL VOC images, producing realistic conditions with heavy raindrops, motion streaks, snowflakes, haze-like artifacts, and reduced visibility. As shown quantitatively in Table~\ref{tab:voc-weather} and qualitatively in Fig.~\ref{fig:snow_rain}. The model demonstrates strong robustness under various synthetic rain and snow degradations, successfully detecting objects despite heavy streaks, dense particles, and visibility loss.

\section{Conclusion}

In this paper, we present C2FXNet, a coarse-to-fine scene expert network for unified object detection across adverse weather. By integrating the Multi-step Reasoning Router, Fine Scene Refinement, and Scene-aware Mixture-of-Experts, our model establishes a coherent text-guided adaptation process that spans coarse recurrent scene reasoning, fine-grained feature modulation, and dynamic expert fusion for robust all-weather detection. Extensive experiments further demonstrate consistently superior accuracy and cross-domain generalization compared with existing methods. However, C2FXNet is trained in a fully supervised manner with limited categories and weather types. Future work will extend it toward open-world all-weather detection that adapts to unseen conditions and novel object categories.

\begin{acks}
This work was supported in part by the National Natural Science Foundation of China under Grants 62462017, 82502451, 82272075, and 62502130, in part by the Natural Science Foundation of Guangxi Province under Grant 2025GXNSFBA069390, in part by the Guangxi Key Research and Development Program under Grant AB2401008, and in part by the Hainan Provincial Natural Science Foundation under Grant 826QN0577.
\end{acks}

\bibliographystyle{ACM-Reference-Format}
\bibliography{ref}

\end{document}